# Performance Study of Low Inertia Magnetorheological Actuators for Kinesthetic Haptic Devices*

Louis-Philippe Lebel, Jean-Alexis Verreault, Jean-Philippe Lucking Bigué, Jean-Sébastien Plante and Alexandre Girard

*Abstract*— **A challenge to high quality virtual reality (VR) simulations is the development of high-fidelity haptic devices that can render a wide range of impedances at both low and high frequencies. To this end, a thorough analytical and experimental assessment of the performance of magnetorheological (MR) actuators is performed and compared to electric motor (EM) actuation. A 2 degrees-of-freedom dynamic model of a kinesthetic haptic device is used to conduct the analytical study comparing the rendering area, rendering bandwidth, gearing and scaling of both technologies. Simulation predictions are corroborated by experimental validation over a wide range of operating conditions. Results show that, for a same output force, MR actuators can render a bandwidth over 52.9% higher than electric motors due to their low inertia. Unlike electric motors, the performance of MR actuators for use in haptic devices are not limited by their output inertia but by their viscous damping, which must be carefully addressed at the design stage.**

## I. INTRODUCTION

Kinesthetic haptic devices provide force feedback during teleoperation tasks or virtual reality (VR) simulations. In either case, haptic devices must render a virtual environment (VE) with a given impedance, such as soft skin or hard bone. From the user's perspective, the rendered impedance must be as realistic as possible to ensure a total immersion. In practice, however, developing high-fidelity haptic devices is challenging since such devices must have (1) a wide rendering area as well as (2) a high rendering bandwidth.

As shown in Fig1, the rendering area (light blue) of a haptic device is defined as the range of impedances that a device can render. This rendering area is lower bound by the passive impedance (red line) of the device and upper bound by the controller stability limit and force capacity (orange line).

A haptic system's passive impedance is critical to render low impedances, and to allow the user to feel completely free when moving in empty space [1]. Mechanical properties of a haptic device, such as inertia and damping, are responsible for its passive impedance, and thus must be minimized by design. On the other hand, a haptic system's high impedance rendering is limited by the maximal force of the device, to avoid force saturation [1], and the controller stability. Z-width, M-width and virtual wall stability criteria are different metrics that have been proposed to quantify the maximum rendered impedances at a given operating frequency[2][3] [4].

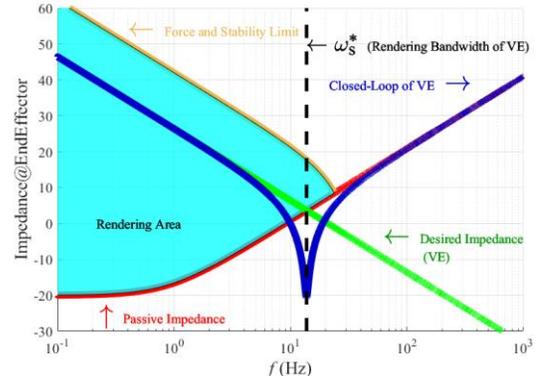

*Fig. 1 : Rendering area of a haptic device and rendering bandwidth under closed-loop rendering of a virtual spring, the desired Virtual Environment (VE)*

To evaluate the rendering quality over a range of frequencies, the effective impedance concept has been proposed. This framework decomposes the overall system impedance into an *effective stiffness*, an *effective damping* and an *effective mass* across its usable frequency domain[5]. As shown in Fig. 1, in the effective impedance framework, a VE (green line) consisting of a virtual spring can be rendered up to a rendering bandwidth, $\omega^*_s$, defined by:

$$\omega^*_s = \sqrt{\frac{K}{m}} \qquad (1)$$

where $K$ is the stiffness of the virtual spring and $m$ the effective mass of the device. The rendering bandwidth of haptic devices must be as high as possible for user to distinguish hard surfaces from one another [6] by stimulating mechanoreceptors of the human body [7]. Beyond $\omega^*_s$, inertial forces dominate the response.

From (1), the rendering bandwidth can only be increased by reducing the mass parameter, $m$, since the stiffness parameter, $K$, is imposed. The effective mass of a haptic device comprises the linkage inertia and the reflected inertia of the actuator [8]. While the linkage inertia is imposed by the device, the actuator inertia can be minimized using low actuator technologies.

Low-impedance kinesthetic haptic devices mainly rely on electric motor (EM) as the primary source of actuation. However, high torque EMs can be bulky, and make use of appropriate gearing to reduce weight. However, a

*Research supported by Natural Sciences and Engineering Concil of Canada (NSERC) and Fond Québécois de la Recherche sur la Nature et les Technologies (FRQNT).

Authors are with is with the Sherbrooke's University

(e-mail: {louis-philippe.lebel2, jean-alexis.verreault, jean-sebastien.plante, alexandre.girard2} @ usherbrooke.ca). and Exonetik (email: {jp.bigue, js.plante} @ exonetik.com) combination of EM and a gear reduction will significantly increase apparent inertia and friction, thus increasing the passive impedance of the device [9]. A popular alternative is the use of Series Elastic Actuators (SEA), where a spring element is placed between the actuator and load to lower the passive impedance. However, the added compliance lowers the system resonance [10], thereby limiting the maximum rendering bandwidth. Admittance system have also shown to have limited rendering bandwidth compared to impedance system [4].

A potential solution to increase both the rendering area and rendering bandwidth simultaneously, is to use magnetorheological (MR) actuators. An MR actuator is composed of a motor (power source) and at least one MR clutch, that is used to actively modulate the torque generated by the motor. MR clutches have better torque-to-inertia ratio than direct drive electric motor and have very high force bandwidth [11]. Thus, MR actuation has the potential to boost the rendering bandwidth of haptic devices for a given virtual environment by reducing the total mass (inertia) at the effector. Previous work studied the use of MR actuator for haptics devices [12]. While this exploratory work suggested promising capabilities for virtual wall rendering, the work did not provide quantitative data on the dynamic performance of MR actuators.

This paper thus presents an analytical and experimental study of the bandwidth area of MR actuators and the rendering bandwidth of virtual spring over its full operational frequency spectrum. Results are compared against equivalent electric motor in order to assess the potential of the MR technology as an actuation alternative in haptics.

## II. MECHANICAL DESIGN

### A. Magnetorheological Actuator

Magnetorheological fluid (MRF) is composed of a carrier fluid and up to 40% of iron particles (3-10 microns)[13] making MRF responsive to magnetic fields. When a magnetic field is applied to the fluid, its yield stress increases in milliseconds, while its viscosity remains unchanged. The principle is used to finely control the torque transferred in MR clutches.

MR actuators overcome the limitation of EM actuator where facing an inevitable trade-off between force and inertia [11]. MR clutches are composed of thin, lightweight shear interfaces (generally drums or disks) minimizing the rotating inertia compared to EM rotor.

The specific clutch design used in this study is shown in Fig. 2. A welded and brazed construction is used to efficiently guide the magnetic flux across the drums. The output rotor is machined from a rough blank made of brazed stainless (rotor) and steel (drums) while the input assembly is made of a welded stainless steel and steel. Electromagnet consists of 140 turns coil of epoxy bonded 28 AWG copper.

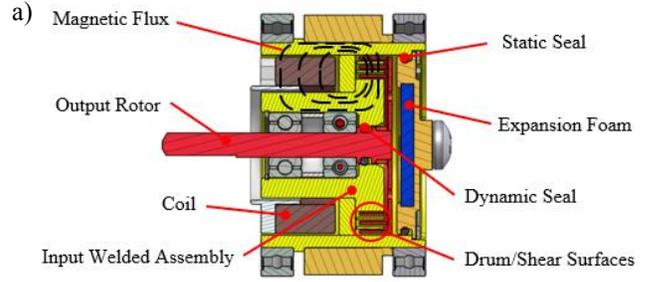
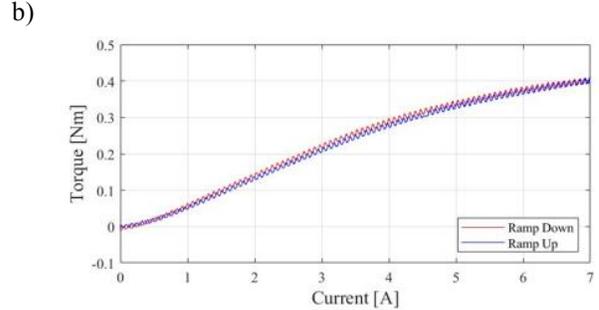

*Fig. 2 : a) Cross section of the test MR clutch; b) Experimental torque vs current loop of the test clutch*

The clutch's blocked output torque is measured experimentally with the test bench described in section B. A linear current ramp up (0 to 7A) of a duration of 10 seconds followed by an identical ramp down is induced in the coil. Input of the clutch is turning at a constant speed of 530 RPM. Results are shown in Fig. 2b and MR clutch specifications are listed in TABLE 1.

The torque versus current loop of Fig. 2b shows negligible magnetic hysteresis. Indeed, a quasi-linear behavior can be observed between 1 and 4 amperes, making MR clutches easily controllable. Above 4 amperes, the magnetic frame begins to saturate, limiting the maximal output torque to ~0.4 Nm. Small torque oscillations observed in Fig. 2b are caused by a mechanical runoff between the input and output rotor of the MR. Those oscillations are not detrimental and could be removed with better machining.

### B. Test Bench

A haptic lever test bench, shown in Fig. 3, is designed to analyze and compare the performance of an EM actuated and MR actuated haptic device under various operating conditions. Both actuations technologies can rapidly and easily be coupled or decoupled to the lever using set screws. Thus, the mechanical properties of the lever itself remain the same for all actuation system. A 17 bits encoder, *Netzer DS-25*, is mounted directly on the lever shaft and is used as position sensor of the haptic lever. High resolution position sensor is used to prevent instability at high rendered stiffness [3]. A 6 axis loadcell, *ATI nano 17*, is mounted at the end-effector of the lever for data acquisition. The lever, composed of a carbon fiber tube, has a length of 170 mm (from rotation axis to center of loadcell) and can be swapped for a shorter 75 mm version to analyze the influence of mechanical gearing on performances.

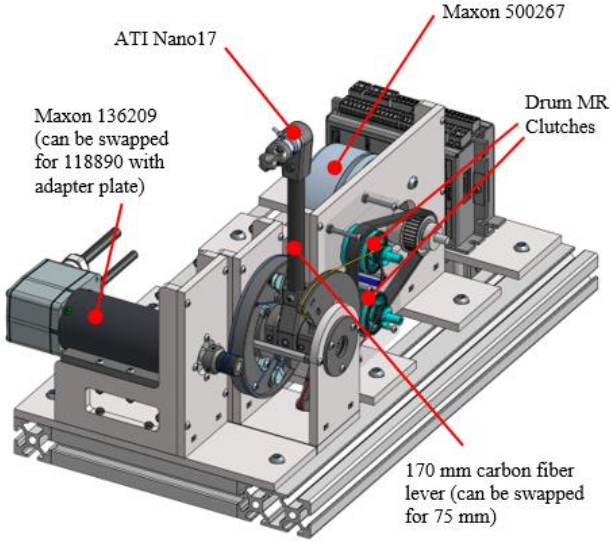

*Fig. 3 : Test bench used to compare the performance of EM and MR actuation technologies. MR clutches support case made transparent for better understanding.*

TABLE 1 : RELEVANT ACTUATORS FOR HAPTIC

| Parameter | Actuators | | | |
|---|---|---|---|---|
| | Proposed drum clutch | Maxon 118890 | Maxon 136206 | Maxon 136209 |
| Rated torque (Nm) | 0.4 | 0.046 | 0.174 | 0.35 |
| Rotor Inertia (g*cm²) | 2.7 | 20 | 119 | 209 |
| Torque/Inertia (N.m/ g*cm²) | 1.48e-1 | 2.30e-3 | 1.46e-3 | 1.67e-3 |
| Torque Density (N.m/ g) | 4.71e-3 | 1.70e-4 | 2.05e-4 | 3.18e-4 |
| Mass (g) | 85 | 270 | 850 | 1100 |
| Overall Dimension (mm) | 30x30x40 | 32x32x60 | 45x50x111 | 45x50x145 |

The MR actuator subsystem uses a pancake style motor (*Maxon 500267*) to drive a pair of 0.4 Nm MR clutches. The motor is speed controlled using a servodrive *ECSON 70/10* to provide a velocity higher than the required output velocity of the MR clutch. An antagonist configuration is selected for the 2 MR clutches, each pulling the lever in its own direction. Cable preloading is unnecessary since the antagonist configuration maintains a minimal tension in the cables at all time due to the clutches viscous drag.

As for the EM subsystem, two ironless low inertia brushless *Maxon* motor, *136209* and *118890* are used in this study to develop an analytical model. A reference motor *136206*, will be used in simulation to compare with the proposed MR actuators. Specification of the motors are listed in Table 1. A capstan drive is used to transfer EM torque to the haptic lever.

Both actuation subsystems are designed to have the same mechanical gearing. A 14 mm winding drums are used on both actuators and ultra-flexible steel cables are used everywhere.

### III. ANALYTICAL DEVELOPMENT

#### A. Model

A 2 degrees of freedom (DOF), shown in Fig. 4b, is proposed to analyze the rendering bandwidth and rendering area of MR and EM system. The actuator, modelled as $m_1$ and $b_1$, produces a force, $F_a$, that is transmitted by the transmission, $k$ and $b$, to move the linkage, $m_2$. Human interaction is integrated to the model with force $F_h$. Because a change of magnetic field doesn't affect the MR fluid viscosity, actuator's damping, $b_1$, can be considered constant for both actuators.

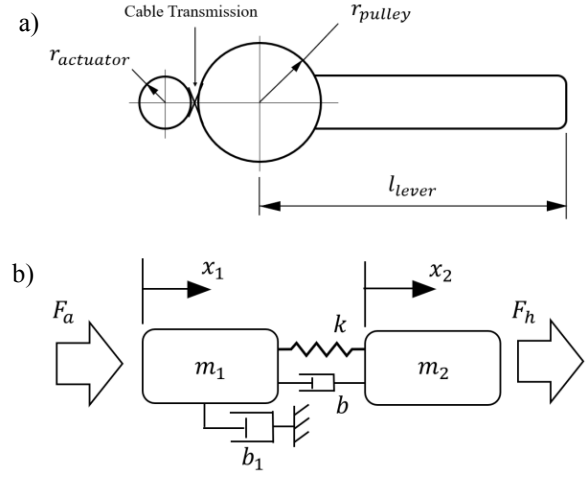

*Fig. 4 : a) System diagram; b) 2-DOF dynamic model.*

The system's 2-DOF impedance is represented by the transfer function:

$$Z_h(s) = \frac{F_h(s)}{\dot{x}_2(s)} = \frac{m_1 m_2 s^3 + (m_2 b_1 + m_1 b + m_2 b)s^2 + (b_1 b + m_1 k + m_2 k)s + b_1 k}{m_1 s^2 + (b_1 + b)s + k} \quad (2)$$

expressed at the lever output in linear coordinates. The rotary to linear conversion is done by considering the system's geometry linking the actuator inertia to its equivalent linear mass:

$$m_1 = I_{actuator} * \left(\frac{r_{pulley}}{r_{actuator}}\right)^2 * \frac{1}{l_{lever}^2} \quad (3)$$

where $r_{pulley}$ is the radius of the lever's pulley, $r_{actuator}$ is the radius of the actuator's pulley, $l_{lever}$ the lever's length and $I_{actuator}$ the actuator's inertia. All parameter are shown in Fig. 4a.

#### B. Parameter Characterization

Parameters of the 2-DOF model are identified with 2 experimental tests.

A first characterization test is used to identify the actuator and transmission properties ($m_1$, $b_1$, $k$ and $b$) by blocking the effector of the lever (Blocking $m_2$). A logarithmic chirp, ranging from 0.1 to 500 Hz, is sent as actuator command and output force is measured.

A second test is used to identify the properties of the lever, that is its equivalent linear mass ($m_2$), by adding a known compliance at the end effector. Again, a 0.1 to 500 Hz chirp excited a broad range of frequencies.

In both characterization tests, identified values, shown in TABLE 2, are manually adjusted from theorical values to a best visual fit the system experimental response. All parameters are "linear" values, that are equivalent to rotary values at end-effector.

TABLE 2: IDENTIFIED MODEL PARAMETERS OF THE 2-DOF MODEL OF THE TEST BENCH WITH THE 170 MM HAPTIC LEVER

| Param. | ACTUATORS | | | | Comments |
|---|---|---|---|---|---|
| | EM system (118890-136209) | | MR system | | |
| | Identified | Theorical | Identified | Theorical | |
| $m_1$ (g) | 3,7-30,8 | 3,7-38,5 | 1,5 | 1,0 | Eq. mass of actuation at end-effector |
| $b_1$ (Ns/m) | 0,1-1,1 | N/A | 1,0 | N/A | Eq. damping of actuator at end-effector |
| $m_2$ (g) | 14,0 | 11,0 | 14,0 | 11,0 | Eq. mass of lever inertia at end-effector |
| $k$ (N/m) | 6 000 | N/A | 4 000 | N/A | Eq. Transmission stiffness |
| $b$ (Ns/m) | 10 | N/A | 20 | N/A | Eq. Transmission damping |

## IV. RESULTS

### A. Experimental Results

The closed-loop impedance performance of each actuation system is experimentally analyzed and compared to model predictions.

Experimental bode and effective impedance are respectfully shown in Fig. 5 for *170 mm* lever and Fig. 6 for *75 mm* lever. Impedance of the end-effector is then decomposed into its main components, effective stiffness, effective damping and effective mass as proposed in [5] that can be analyzed through their frequency domain.

Data are obtained by stimulating a broad range of frequencies at the end-effector. Low frequencies are manually excited by moving the end-effector of the haptic lever and higher frequency are excited by providing impacts on the end effector. At least 3 minutes of data are used to plot the end-effector impedance, equally alternating between manual and impact excitation. 2 levers length, 170mm and 75 mm, are used to analyze impact of mechanical advantages.

The rendering bandwidth is affected by the total reflected inertia of the system, as discussed in section III. Thus, by changing the lever length to from *170 mm* to *75 mm*, the reflected inertia of actuators is increased by the square of the length change changing the total mass of the system. Performance of the high inertia actuator (136209) is significantly reduced by such a ratio change, with a rendering bandwidth nearly cut in half as demonstrated in Fig. 6.

Fig. 5 and Fig. 6 show that model predictions using the parameters of TABLE 2 are in good agreement with experimental data over widely varying operating conditions consisting of a 0-100 Hz frequency spectrum, 3 different actuator sizings, and 2 different mechanical gearings. Model fidelity and accuracy, as limited to the boundaries of this work, are thus confirmed.

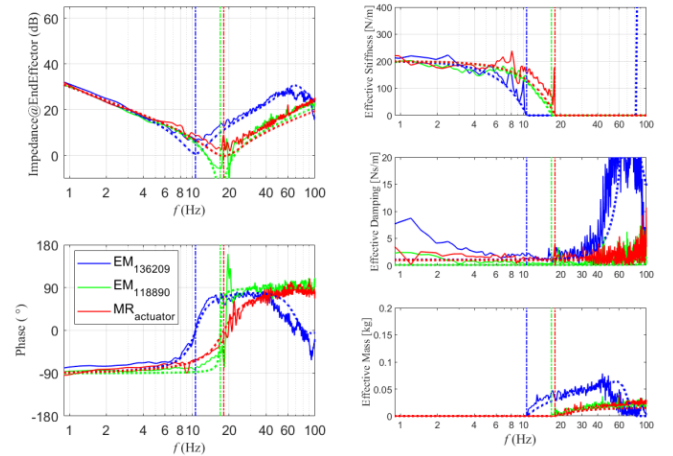

*Fig. 5 : Experimental bode and effective impedance plot of both actuation system rendering a 200 N/m stiffness with 170 mm lever. Full lines are experimental data and dotted lines are the corresponding model predictions.*

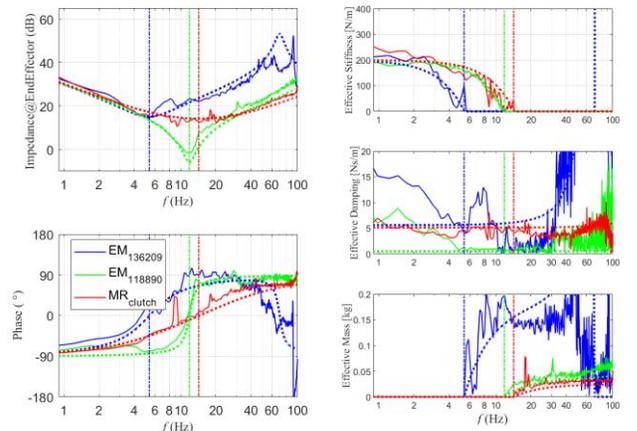

*Fig. 6: Experimental bode and effective impedance plot of both actuation system rendering a 200 N/m stiffness with 75 mm lever. Full lines are experimental data and dotted lines are the corresponding model predictions.*

### B. Simulation Extrapolation

The rendering area of equivalent EM and MR actuated systems are shown in Fig. 7. Rendering areas are created by filling the area between the simulated passive impedance and closed-loop performance predictions (as proposed in [5]) of each system when rendering a given VE taking the form of a maximum stiffness $K_{max}$ placed as a bi-lateral centering spring. $K_{max}$ is defined as the maximal stiffness that can be rendered in a given workspace, $d$, without force saturation using the following equation:

$$K_{max} = \frac{T_{max} * \left(\frac{r_{pulley}}{r_{actuator}}\right) * \left(\frac{1}{l_{lever}}\right)}{d} \qquad (4)$$

where $T_{max}$ is the maximum torque of the actuator. A perfect comparison between EM and MR regarding $T_{max}$ is difficult since EM maximal force output is limited by their thermal behavior while MR clutches maximal torque is limited by magnetic saturation and thermal dissipation due to slip in the MR clutch. Thus, authors have fixed, to the best of their knowledge, the maximal torque criteria between EM and MR systems. $T_{max}$ is determined for EM using manufacturer design guide lines [14] with a 25% duty cycle considering a one minute cycle. $T_{max}$ is thus taken as twice the value of the rated torque of TABLE 1 for EM. Regardless of $T_{max}$ values, tendencies observed in simulations reveal main trends.

Other simulation parameters include a 1 *ms* pure delay for MR actuators, corroborating experiments and previous results for similar designs [15]. Discretization and sampling time of all systems are considered infinite for simplicity. Finally, another electric motor, *Maxon 136206* with 0.174 Nm rated torque, is introduced to extend the comparison. *Maxon 136206* properties are defined with manufacturer specifications.

Fig. 7 show the effects of various design parameters on rendering areas and rendering bandwidths. When actuators are actively rendering an impedance, the closed-loop impedances as defined by $K_{max}$ show 2 resonances. The first one is the rendering bandwidth $\omega^*_s$ anti-resonance (Fig. 7c) where the virtual spring ($K_{max}$) resonates with the two masses ($m_1, m_2$) moving in solid body motion with the transmission spring ($k$) not deforming. The second one is the system's resonance where the transmission spring resonates with the two masses not in solid body motion (Fig. 7b,c). When actuators are turned off and the system is backdriven, the passive impedance curves only show the system resonance. All systems tend to the lever's impedance after their respective system resonances.

Fig. 7c shows an actuator performance comparison for a same force output. Fixing a design parameter is necessary for a fair comparison of haptic performance of different technologies. Thus, EM and MR actuators are selected to have the same maximal outputted force as a case study. As shown in Fig. 7c, passive impedance of EM and MR system are differently composed with EM having more mass and MR more damping but have a small contribution to the reflected inertia. Rendering bandwidth of $K_{max}$ can be compared, where MR system has 52.8% more rendering bandwidth than an equal output force EM. However, MR system has ~66% more damping than equivalent EM system, limiting the rendering area.

Actuator gearing and/or scaling is often used to boost output forces. The effect of gearing and scaling on dynamic performance on both EM and MR technologies is exposed on Fig. 7a and b.

Fig. 7a shows the effect of actuator gearing on performance. Gearing a MR clutch has a limited impact on overall reflected inertia. 4X gearing increase by 145% the total inertia of the MR system. The effect of 4X gearing increase damping of both system by 1600%, since the grounded damper $b_1$ comes from the actuator. Initial damping of MR is higher than EM, making 4X geared MR system highly damped. With respect to gearing, each technology is bounded, but not by the same physical phenomenon.

Finally, Fig. 7b shows the effect of actuator scaling on performance. Scaling a MR clutch has very low impact on overall reflected inertia of the system. 2X scaling increase by 14% the total reflected inertia of MR system. As observed in Fig. 7b, scaling EM increase rapidly total reflected inertia of the system. Scaling is an overall best option for both EM and MR compared to gearing due to less passive impedance increase in general. However, size or weight requirement could limit the scaling possibilities.

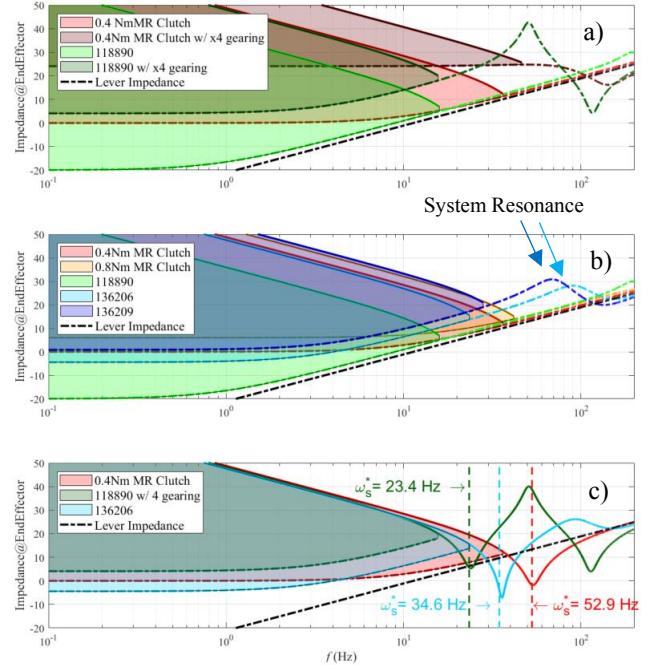

*Fig. 7 : a) Effect of gearing on MR and EM actuation on rendering bandwidth; b) Effect of actuator scaling on rendering bandwidth; c) Comparison of rendering area for a given max force with identified rendering bandwidth of $K_{max}$. Full lines are closed-loop impedance and dashed lines passive impedance.*

Scaling the 136206 EM to 136209 increases actuator reflected inertia thereby lowering the system resonance as observed in Fig. 7b. Thus, low inertia actuators in haptics have the potential to have high mechanical bandwidth [11] or tolerate low stiffness transmission while keeping good dynamic performance.

V. DISCUSSION

Both viscous damping and inertia are proportional to the square of the gearing ratio, making the passive impedances of both systems sensible to gearing. As observed in section IV, the passive impedance of MR system has more damping than EM system and EM system has more reflected inertia than MR system for a same force output capability.

Regarding inertia, systems where actuator inertia is non-negligible compared to the linkage, such as EM 136209, will have considerable increase of reflected inertia when gearing increases, and thus a significant reduction in their rendering bandwidth.

Regarding damping, the total damping will increase rapidly for geared highly damped systems such as MR actuation,

increasing free moving forces. Here, software compensation can be considered to reduce passive impedance, by reducing friction. Software compensation such as natural admittance can ensure stability when reducing friction, but can't reduce high level of physical inertia [16]. Furthermore, MR clutches generate a highly linear damping response across the frequency domain, as observed in Fig. 5 and Fig. 6, making it easier for software compensation, even for highly geared system. However, EM actuated systems will be limited by inertia, that is hard to compensate.

Finally, the performance assessment is completed by looking at each technology weights by considering a generic 3 DOF haptic system. In separate works, the test MR clutches used in this work were integrated into a 3-DOF actuation module, see Fig. 8. The module contains 6 MR clutches, one *Faulhaber 4221 motor*. The complete module weighs ~1000gr with all components and frame. Using the numbers in Table 1, an equivalent 3 DOF EM counterpart using the closest performing motor (*Maxon 136206*) would weighs 2550 gr for the motors only (no structure or transmission). The MR technology could significantly reduce weights by being about ~2.5X lighter.

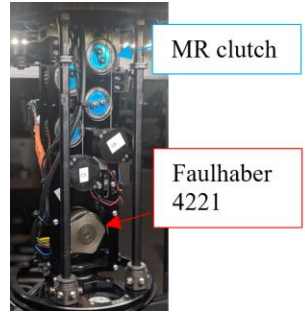

*Fig. 8: 3-DOF MR module*

## VI. CONCLUSION

This paper addresses and analyze the effect of EM and MR actuators properties in haptic devices and quantified their effect on rendering area and rendering bandwidth of a virtual spring. An analytical model is presented and validated using a haptic lever as a test bench. A drum MR clutch designed to have a good torque-to-inertia ratio is compared to an equivalent motor. Results show that actuators inertia has an influence on overall performance of a haptic device, especially for geared system. For a same output forces, VE rendering bandwidth can be increased (52.9%) with low inertia actuator such as MR actuators. Simulation have shown that low inertia actuator increases rendering bandwidth by reducing overall reflected inertia leading to more realistic VE [4]. A 3-DOF module also shown weight reduction of ~2.5X lighter than 3 force equivalent EM. However, the rendering area of MR clutches are limited by their highly damped passive impedance nature.

Future work will explore new MR clutches design and control methods to further reduce passive impedance of MR actuation. Lower inertia can be obtained by reducing drum thickness or using other architectures (e.g.: disk vs. drums). Furthermore, improved clutch design could significantly reduce overall damping by widening the shear interfaces gaps, using a longer but lower diameter clutch, or by using low viscosity fluids.